\documentclass[letterpaper]{article} 
\usepackage{aaai2026}  
\usepackage{times}  
\usepackage{helvet}  
\usepackage{courier}  
\usepackage[hyphens]{url}  
\usepackage{graphicx} 
\usepackage{natbib}  
\usepackage{caption} 
\usepackage{algorithm}
\usepackage{algorithmic}
\usepackage{booktabs}
\usepackage{multirow}
\usepackage{subcaption}
\usepackage{makecell} 

\usepackage{newfloat}
\usepackage{listings}
\DeclareCaptionStyle{ruled}{labelfont=normalfont,labelsep=colon,strut=off} 
\floatstyle{ruled}
\newfloat{listing}{tb}{lst}{}
\floatname{listing}{Listing}
\usepackage{amsmath}
\usepackage{amssymb}
\title{HGATSolver: A Heterogeneous Graph Attention Solver for Fluid–Structure Interaction}
\author{
    Qin-Yi Zhang\textsuperscript{\rm 1, 2}\equalcontrib,
    Hong Wang\textsuperscript{\rm 3}\equalcontrib,
    Siyao Liu\textsuperscript{\rm 4},
    Haichuan Lin\textsuperscript{\rm 1, 2},
    Linying Cao\textsuperscript{\rm 1, 2},
    Xiao-Hu Zhou\textsuperscript{\rm 1, 2},
    Chen Chen\textsuperscript{\rm 1, 2},
    Shuangyi Wang\textsuperscript{\rm 1, 2}\thanks{Corresponding authors.},
    Zeng-Guang Hou\textsuperscript{\rm 1, 2}\footnotemark[2]
}
\affiliations{
    \textsuperscript{\rm 1}State Key Laboratory of Multimodal Artificial Intelligence Systems, Institute of Automation, Chinese Academy of Sciences, Beijing 100190, China\\
    \textsuperscript{\rm 2}School of Artificial Intelligence, University of Chinese Academy of Sciences, Beijing 100049, China\\
    \textsuperscript{\rm 3}University of Science and Technology of China, Hefei 230026, China\\
    \textsuperscript{\rm 4}Chengdu University of Technology, Chengdu 610059, China\\
    \{zhangqinyi2024, shuangyi.wang, zengguang.hou\}@ia.ac.cn
}

\begin{document}

\maketitle
\begin{abstract}
Fluid–structure interaction (FSI) systems involve distinct physical domains, fluid and solid, governed by different partial differential equations and coupled at a dynamic interface. While learning-based solvers offer a promising alternative to costly numerical simulations, existing methods struggle to capture the heterogeneous dynamics of FSI within a unified framework. This challenge is further exacerbated by inconsistencies in response across domains due to interface coupling and by disparities in learning difficulty across fluid and solid regions, leading to instability during prediction. To address these challenges, we propose the Heterogeneous Graph Attention Solver (\textbf{HGATSolver}). HGATSolver encodes the system as a heterogeneous graph, embedding physical structure directly into the model via distinct node and edge types for fluid, solid, and interface regions. This enables specialized message-passing mechanisms tailored to each physical domain. To stabilize explicit time stepping, we introduce a novel physics-conditioned gating mechanism that serves as a learnable, adaptive relaxation factor. Furthermore, an Inter-domain Gradient-Balancing Loss dynamically balances the optimization objectives across domains based on predictive uncertainty. Extensive experiments on two constructed FSI benchmarks and a public dataset demonstrate that HGATSolver achieves state-of-the-art performance, establishing an effective framework for surrogate modeling of coupled multi-physics systems.
\end{abstract}

\begin{links}
    \link{Code}{https://github.com/Qin-Yi-Zhang/HGATSolver}
\end{links}

\section{Introduction}

Numerical simulation of coupled multi-physics systems, governed by interacting sets of partial differential equations (PDEs), is a formidable challenge in computational science~\cite{dowell2001modeling}. Fluid–Structure Interaction (FSI) is a prime example, with applications ranging from aircraft design~\cite{kamakoti2004fluid} to cardiovascular hemodynamics~\cite{singh2024hemodynamic,zhang2024learning}. The crux lies not in discretizing the fluid or solid domains separately but in enforcing kinematic and dynamic continuity at their complex, moving interface~\cite{hou2012numerical}. Traditional solvers face stability issues and prohibitive computational costs when handling this coupling, particularly in regimes with strong added-mass effects~\cite{wiggert2001fluid}, highlighting the need for efficient, learning-based surrogate solvers~\cite{azizzadenesheli2024neural, luo2024neural, wang2025symmap, wangaccelerating, dong2024accelerating}.
\begin{figure}[t]
\centering
\includegraphics[width=1.0\columnwidth]{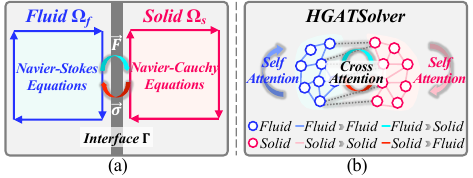}
\caption{Conceptual overview of HGATSolver. (a) The FSI system couples fluid and solid domains with distinct PDEs and a dynamic interface. (b) We encode this structure as a heterogeneous graph, enabling type-aware attention for intra- and inter-domain physics.}
\label{fig:concept}
\end{figure}

Neural operators, which learn PDE solution mappings, have emerged as a powerful paradigm~\cite{azizzadenesheli2024neural}. The Fourier Neural Operator delivers remarkable efficiency on regular, structured grids~\cite{li2021fourier}. To accommodate the irregular meshes characteristic of complex geometries, recent work has turned to graph-based and attention-driven architectures capable of operating on arbitrary discretizations~\cite{NEURIPS2023_70518ea4, pmlr-v235-wu24r, gao2025discretizationinvariance}. 

Despite these advances, when applied to coupled systems like FSI, the prevailing approach of consolidating the entire system into a single, homogeneous graph introduces a fundamental mismatch in the model architecture. This unified approach forces a universal message-passing scheme to approximate distinct physical laws, such as the Navier-Stokes and elastodynamic equations~\cite{coutand2006interaction}, within the same framework. Consequently, the model fails to exploit the system’s known physical decomposition as a structural inductive bias. As a result, the model must expend considerable resources to rediscover this separation, thereby increasing the complexity of the search space. This limits its physical consistency and generalizability.

Beyond this structural limitation, the learning-based FSI solver confronts two inherent difficulties of the underlying physics. First, the strong coupling at the fluid-solid interface often results in a numerically stiff system, making explicit time-stepping schemes prone to instability. An effective solver requires a mechanism to ensure stable predictions, especially under strongly coupled dynamics across the interface. Second, the training process itself presents a challenge in balancing multiple objectives. The model must minimize prediction errors for both the fluid and solid domains, whose governing equations can differ in scale and sensitivity. Manually tuning the weight of each component during training is fragile and suboptimal.
\begin{figure}[t]
  \centering
  \includegraphics[width=1.0\columnwidth]{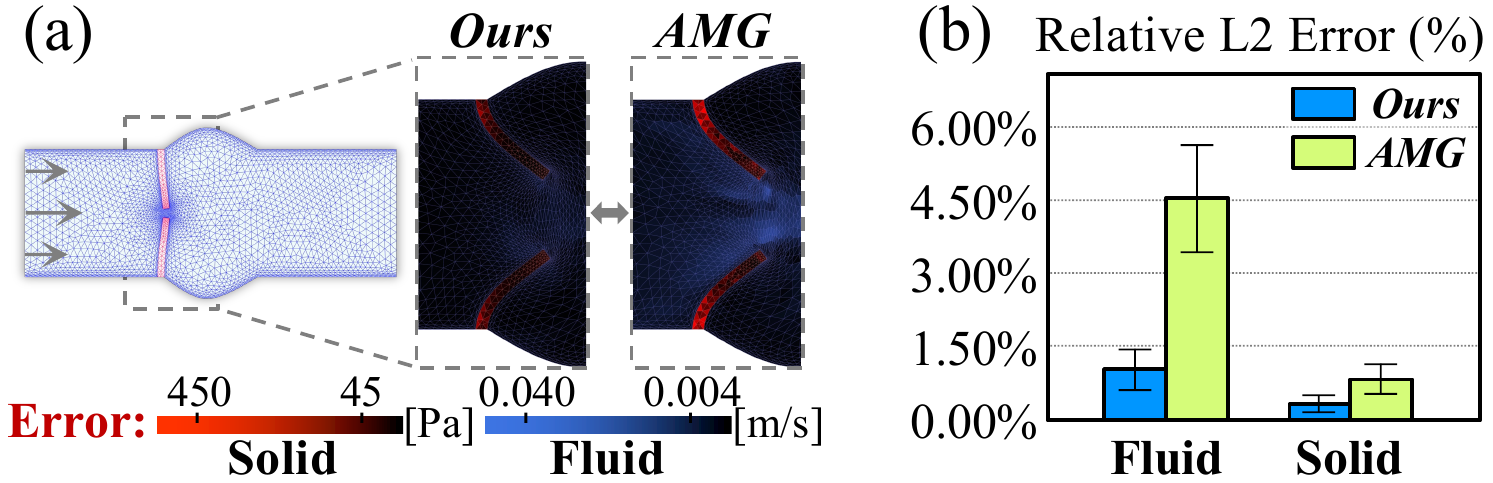}
    \caption{Comparison in the interaction region on FI-Valve. (a) Enlarged error maps show reduced error near the interface. (b) Relative $\ell_2$ errors for fluid and solid domains demonstrate HGATSolver’s superior accuracy compared to a GAT-based baseline.}
  \label{fig:Fig2}
\end{figure}

To address these challenges, we introduce the Heterogeneous Graph Attention Solver (HGATSolver), a framework built on three core principles. First, to resolve the architectural mismatch, we represent the system using a heterogeneous graph~\cite{wang2022survey}, as shown in Fig.~\ref{fig:concept}. This graph assigns distinct node types to the fluid and solid domains and uses typed edges to capture both the internal dynamics of each domain and the coupling conditions at the interface~\cite{zhao2021heterogeneous}. This representation directly embeds the system’s physical decomposition into the model. As illustrated in Fig.~\ref{fig:Fig2}, it results in lower prediction errors compared to homogeneous baselines. Second, to ensure numerical stability, we introduce a Physics-Conditioned Gating Mechanism (PCGM). This component adapts the state update by acting as a learned, state-dependent relaxation factor, preventing instabilities in the simulation. Third, to achieve balanced and robust training, we propose Inter-domain Gradient-Balancing Loss (IGBL), allowing the model to autonomously weight the loss for each physical domain based on its learned aleatoric uncertainty~\cite{kendall2018multi}. This eliminates the need for manual loss weighting and enhances the robustness of the training process.

Our work makes the following contributions:

\begin{itemize}
\item We propose the first FSI simulation framework built on a heterogeneous graph architecture, which encodes strong physical inductive biases by explicitly encoding the system's distinct domains and their interface.

\item We propose a robust learning framework for stiff, coupled systems, combining PCGM for stability and IGBL for uncertainty-aware training across domains.

\item  We construct two challenging FSI benchmarks featuring distinct geometries and physical boundary conditions. The first benchmark focuses on flow-induced structural deformation under high-Reynolds-number conditions, while the second models structure-induced fluid dynamics with strong loading. HGATSolver achieves state-of-the-art performance on these benchmarks and a public dataset.
\end{itemize}

\section{Related Works}
\subsection{Learning-based PDE Solvers}
Learning-based solvers have advanced from the grid-dependent FNO~\cite{li2021fourier} to graph-based operators that accommodate the unstructured meshes inherent to complex geometries~\cite{NEURIPS2023_70518ea4}. Token-wise attention is theoretically shown to approximate the PDE solution operator as a learnable integral operator~\cite{hao2023gnot, pmlr-v235-wu24r, wangmixture, huang2025self}. However, prevailing approaches for single-physics problems adopt a structurally uniform graph representation, disregarding domain-specific heterogeneity~\cite{AMG2025, liu2025aerogto}. This limitation persists in multi-physics pre-trained models like CoDA-NO~\cite{rahman2024pretraining}, which distinguish variables via token types but do not encode the structural separation of physical subdomains. As a result, these models must infer domain boundaries from data rather than leveraging them as structural priors~\cite{wang2025stnet, lv2025exploiting, wei2025mecot, wei2025vflow}.

\subsection{Deep Learning for FSI}
FSI presents unique challenges for deep learning due to the coupling of fluid and solid physics through moving interfaces. These systems are often stiff and exhibit highly heterogeneous behaviors, complicating accurate simulation~\cite{zhu2023physics, gao_pinn}.
Prevalent approaches employ CNNs on structured grids~\cite{han2022deep, hu2024fast}, but suffer from mesh distortion issues. Recent methods leverage graph-based models to support unstructured meshes~\cite{gao2024predicting, fan2024differentiable}, yet most assume uniform node and edge types, overlooking the distinct physical laws governing fluid and solid regions~\cite{xu2024deep}.
Training is further complicated by the need to learn dynamics across domains with differing numerical scales jointly. Manually tuning fixed loss weights often leads to instability or suboptimal convergence~\cite{liu2024novel, bublik2023neural}. These limitations highlight the need for models that explicitly encode physical heterogeneity and adaptively learn inter-domain interactions.

\subsection{Heterogeneous Graph Neural Networks}
Research in Heterogeneous Graph Neural Networks has established two primary strategies: those reliant on predefined meta-paths to capture semantics~\cite{fu2020magnn, wang2019heterogeneous}, and more flexible meta-relation approaches that learn type-based interactions directly~\cite{hu2020heterogeneous, wang2022survey}. The fundamental structure of a coupled multi-physics system—comprising distinct domains governed by unique physical laws and linked by well-defined interface conditions—presents a natural and compelling mapping to the heterogeneous graph formalism~\cite{bublik2023neural, liu2024novel, yang2023simple}. While this representation provides a powerful structural prior, modeling the dynamic interactions across different edge types poses a considerable challenge for ensuring stable and accurate prediction.

\section{Preliminaries}
\label{sec:preliminaries}
This section defines the FSI governing physics, links neural operators to graph attention, and introduces heterogeneous graphs as a structural prior.
\subsection{Governing Equations of FSI}
An FSI system comprises a fluid domain $\Omega_f$ and a solid domain $\Omega_s$ coupled at a moving interface $\Gamma_i$. Coupled PDEs govern the system's evolution.

Within $\Omega_f$, the dynamics of an incompressible fluid are described by the Navier-Stokes equations:
\begin{equation}
\label{eq:navier_stokes}
\rho_f \left( \frac{\partial \mathbf{u}_f}{\partial t} + \mathbf{u}_f \cdot \nabla \mathbf{u}_f \right) = \nabla \cdot \boldsymbol{\sigma}_f, \quad \nabla \cdot \mathbf{u}_f = 0,
\end{equation}
where $\rho_f$ is the fluid density, $\mathbf{u}_f$ is the velocity, and $\boldsymbol{\sigma}_f = -p\mathbf{I} + \mu_f(\nabla \mathbf{u}_f + (\nabla \mathbf{u}_f)^T)$ is the fluid stress tensor. In the solid domain $\Omega_s$, the elastodynamics are governed by the Navier-Cauchy equations:
\begin{equation}
\label{eq:navier_cauchy}
\rho_s \frac{\partial^2 \mathbf{d}_s}{\partial t^2} = \nabla \cdot \boldsymbol{\sigma}_s,
\end{equation}
where $\rho_s$ is the solid density, $\mathbf{d}_s$ is the displacement, and $\boldsymbol{\sigma}_s$ is the solid stress tensor. The two physics are coupled at the interface $\Gamma_i$ by kinematic (no-slip) and dynamic (force equilibrium) conditions:
\begin{align}
\label{eq:kinematic_constraint}
\mathbf{u}_f &= \frac{\partial \mathbf{d}_s}{\partial t} && \text{on } \Gamma_i, \\
\label{eq:dynamic_constraint}
\boldsymbol{\sigma}_f \cdot \mathbf{n} &= \boldsymbol{\sigma}_s \cdot \mathbf{n} && \text{on } \Gamma_i.
\end{align}
Here, $\mathbf{n}$ is the outward unit normal vector to the interface $\Gamma_i$. The core challenge lies in robustly enforcing these coupling conditions at the complex, moving interface.

\subsection{Graph Attention as a Neural Operator}

Neural operators aim to learn mappings between infinite-dimensional function spaces by approximating solution operators as integral transforms~\cite{hao2023gnot,pmlr-v235-wu24r}. Given an input field $u: \Omega \to \mathbb{R}^C$ and a target location $g^* \in \Omega$, the operator can be written as:
\begin{equation}
\label{eq:integral}
G(u)(g^*) = \int_{\Omega} \kappa(g^*, \xi) u(\xi) \, d\xi,
\end{equation}
where $\kappa: \Omega \times \Omega \to \mathbb{R}$ is a kernel function.

On discrete domains, this integral is approximated by a weighted summation over neighbors:
\begin{equation}
\label{eq:discrete}
G(u)(g^*) \approx \sum_{\xi_j \in \mathcal{N}(g^*)} \alpha_{g^*, j} \, u(\xi_j),
\end{equation}
where $\alpha_{g^*, j}$ are normalized attention weights derived from the kernel $\kappa(g^*, \xi_j)$.

This formulation aligns with graph attention mechanisms, where the output at node $i$ is computed as:
\begin{equation}
\label{eq:gat}
\mathbf{h}'_i = \sum_{j \in \mathcal{N}(i)} \alpha_{ij} \, \mathbf{W} \mathbf{h}_j,
\end{equation}
with $\mathbf{h}_j$ denoting node features, $\mathbf{W}$ a learnable projection, and $\alpha_{ij}$ a learnable, data-dependent attention weight.

Thus, graph attention can be viewed as a discretized, learnable approximation of an integral operator over the domain $\Omega$~\cite{AMG2025}.

\subsection{Heterogeneous Graphs as a Physical Inductive Bias}
In FSI systems, fluid and solid regions are governed by different physical laws. Applying a single attention kernel across all regions fails to reflect this structural difference.

To address this, we represent the domain as a heterogeneous graph $\mathcal{G} = (V, E, \mathcal{T}_V, \mathcal{T}_E)$, where each node is assigned a type in $\mathcal{T}_V = {\text{fluid}, \text{solid}}$, and each edge belongs to a relation type in $\mathcal{T}_E = {\text{f2f}, \text{s2s}, \text{f2s}, \text{s2f}}$, representing fluid–fluid, solid–solid, fluid-to-solid, and solid-to-fluid interactions, respectively. Each edge type corresponds to a specific physical interaction, such as internal dynamics in fluid or solid domains or coupling across the interface.

We define a relation-aware message passing operator. For a node $i$, the updated feature is computed as:
\begin{equation}
\label{eq:heterogeneous_mp}
\mathbf{h}'_i = \sum_{\tau \in \mathcal{T}_E} \sum_{j \in \mathcal{N}_i^{(\tau)}} \alpha_{ij}^{(\tau)} \, \mathbf{W}^{(\tau)} \mathbf{h}_j,
\end{equation}
where $\mathcal{N}_i^{(\tau)}$ is the set of neighbors of type $\tau$, $\alpha_{ij}^{(\tau)}$ is a relation-specific attention weight, and $\mathbf{W}^{(\tau)}$ is a learnable projection matrix for each relation.

This formulation allows the model to learn distinct interaction kernels for fluid dynamics, solid deformation, and interface coupling, embedding physical structure directly into the network.

\begin{figure*}[t]
\centering
\includegraphics[width=\textwidth]{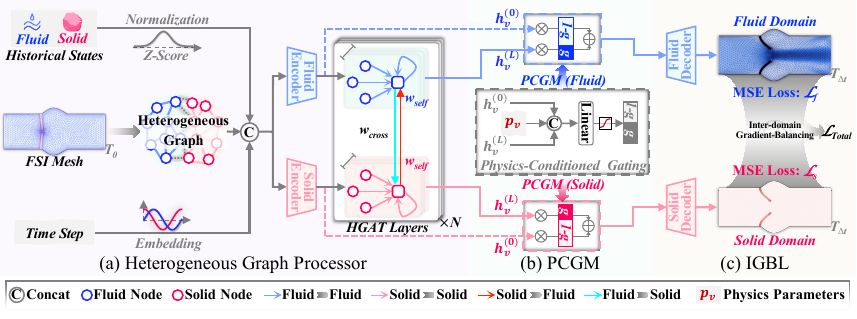}
\caption{Overview of HGATSolver. (a) The FSI mesh is encoded as a heterogeneous graph with fluid and solid nodes and relation-aware edges. (b) PCGM adaptively blends updated and initial states based on physics parameters. (c) IGBL adjusts domain-wise loss weights based on predictive uncertainty.}
\label{fig:architecture}
\end{figure*}

\section{Method}
HGATSolver incorporates three key innovations aligned with the physical nature of FSI: (1) a heterogeneous graph attention processor that distinguishes intra- and inter-domain dynamics; (2) an adaptive gating mechanism that modulates interface coupling based on local and global physical context; and (3) a principled uncertainty-based loss to jointly optimise across heterogeneous domains. An overview is illustrated in Fig.~\ref{fig:architecture}.

\subsection{Heterogeneous Graph Processor}
Let $\mathcal{G} = (V, E)$ denote the heterogeneous graph with node types $\mathcal{T}_V$ and edge types $\mathcal{T}_E$ as defined above. For each node $v \in V$, the input feature vector $\mathbf{x}_v \in \mathbb{R}^{d_{\text{in}}}$ is constructed as:
\[
\mathbf{x}_v = \left[ \mathbf{x}_v^{\text{state}} \,\|\, \mathbf{x}_v^{\text{pos}} \,\|\, \mathbf{x}_v^{\text{time}} \right],
\]
where $\mathbf{x}_v^{\text{state}}$ contains $N$ past frames of normalized physical quantities (e.g., velocity, displacement), $\mathbf{x}_v^{\text{pos}}$ is the spatial coordinate, and $\mathbf{x}_v^{\text{time}}$ is a sinusoidal embedding of the time step $\Delta t$.

This input is encoded via a type-specific MLP:
\begin{equation}
\mathbf{h}_v^{(0)} = \phi_{\text{enc}}^{(\tau_v)}(\mathbf{x}_v), \quad \tau_v \in \mathcal{T}_V.
\end{equation}

We then apply $L$ layers of heterogeneous graph attention, passing messages according to the relation type $\tau \in \mathcal{T}_E$. For each edge $(j \rightarrow i)$ of type $\tau$, we compute the attention energy as:

\begin{equation}
e_{ij}^{(\tau)} = \mathbf{a}^{(\tau)\top} \cdot \sigma_{\text{LeakyReLU}} \left( \mathbf{W}_Q^{(\tau)} \mathbf{h}_i + \mathbf{W}_K^{(\tau)} \mathbf{h}_j \right),
\end{equation}
Here, $\mathbf{W}_Q^{(\tau)}, \mathbf{W}_K^{(\tau)}, \mathbf{W}_V^{(\tau)} \in \mathbb{R}^{d \times d}$ are relation-specific learnable projection matrices, and $\mathbf{a}^{(\tau)} \in \mathbb{R}^d$ is a relation-specific attention vector.
which is then normalized across the neighborhood $\mathcal{N}_i^{(\tau)}$ to obtain the attention coefficient:
\begin{equation}
\alpha_{ij}^{(\tau)} = \frac{\exp(e_{ij}^{(\tau)})}{\sum_{k \in \mathcal{N}_i^{(\tau)}} \exp(e_{ik}^{(\tau)})}.
\end{equation}

Messages are first aggregated per relation type:
\begin{equation}
\mathbf{m}_i^{(\tau)} = \sum_{j \in \mathcal{N}_i^{(\tau)}} \alpha_{ij}^{(\tau)} \cdot \mathbf{W}_V^{(\tau)} \mathbf{h}_j.
\end{equation}

We group edge types into intra-domain ($\mathcal{T}_{\text{self}}$) and inter-domain ($\mathcal{T}_{\text{cross}}$) sets, where $\mathcal{T}_{\text{self}}$ includes edges within the same domain (e.g., f2f, s2s), and $\mathcal{T}_{\text{cross}}$ includes cross-domain types (e.g., f2s, s2f). The total message to node $i$ is:
\begin{equation}
\mathbf{m}_i = w_{\text{self}}^{(\tau_i)} \cdot \sum_{\tau \in \mathcal{T}_{\text{self}}} \mathbf{m}_i^{(\tau)} + w_{\text{cross}}^{(\tau_i)} \cdot \sum_{\tau \in \mathcal{T}_{\text{cross}}} \mathbf{m}_i^{(\tau)},
\end{equation}
with learnable weights $w_{\text{self}}^{(\tau_i)}, w_{\text{cross}}^{(\tau_i)} \in \mathbb{R}_+$ and $\tau_i$ denoting the type of node $i$.

Each layer updates node features via residual connection, normalization, and nonlinearity:
\begin{equation}
\mathbf{h}_i^{(l+1)} = \mathbf{h}_i^{(l)} + \sigma \left( \text{LayerNorm}(\mathbf{m}_i) \right).
\end{equation}

This relation-aware processor enables HGATSolver to learn physically grounded attention kernels specialized for intra-domain dynamics and interface-driven inter-domain coupling.

\subsection{PCGM: Physics-Conditioned Gating Mechanism}

Coupled FSI systems exhibit pronounced disparities in dynamics between fluid and solid regions and discontinuities in response across their shared interface~\cite{augier2015experimental}. These inconsistencies—arising from differences in physical scales, stiffness, and learning difficulty—can lead to numerical instability or overfitting when using fixed, unmoderated GNN updates~\cite{liu2021graph}. To address this, we introduce the \textbf{PCGM}, a learnable, adaptive relaxation factor for each node.

Rather than fully committing to the raw graph-updated state, each node softly interpolates between its initial (pre-message-passing) representation and its updated state, with the interpolation strength determined by local features and global physics parameters.

Let $\mathbf{h}_v^{(0)}$ denote the initial node representation from the encoder, and $\mathbf{h}_{v}^{(L)}$ the output after $L$ layers of heterogeneous message passing. Let $\mathbf{p}_v \in \mathbb{R}^{d_p}$ be the normalized vector of static physics parameters (e.g., material density, fluid viscosity). The gating coefficient $g_v \in (0, 1)$ is computed as:

\begin{equation}
g_v = \sigma_{\text{sigmoid}} \left( \mathbf{W}_g \left[ \mathbf{h}_v^{(0)} \,\|\, \mathbf{h}_{v}^{(L)} \,\|\, \mathbf{p}_v \right] + \mathbf{b}_g \right),
\end{equation}
where $\mathbf{W}_g$ and $\mathbf{b}_g$ are learnable parameters.

The final node representation is then given by:
\begin{equation}
\mathbf{h}_v^{\text{final}} = (1 - g_v) \cdot \mathbf{h}_v^{(0)} + g_v \cdot \mathbf{h}_{v}^{(L)}.
\end{equation}

This mechanism enables each node to learn a domain- and context-specific update strength. Importantly, the PCGM does not directly enforce numerical stability. Instead, it provides a smooth, data-driven control over state evolution, mitigating sharp transitions across the fluid–solid interface and moderating spatiotemporal stiffness. This formulation reflects the relaxation strategies used in classical iterative solvers~\cite{van2011partitioned}, but enables them to be learned end-to-end within a neural framework.

\subsection{IGBL: Inter-domain Gradient-Balancing Loss}

FSI involves two physically distinct yet tightly coupled domains, each governed by different dynamics, numerical scales, and learning challenges. Jointly optimizing their predictions under a unified loss is often unstable, as fixed loss weightings fail to adapt to inter-domain discrepancies, especially near the interface.

Inspired by uncertainty-based task weighting~\cite{kendall2018multi}, we treat predictions in each domain as samples from a Gaussian distribution with domain-specific homoscedastic variance. For each domain $d \in \{\text{fluid}, \text{solid}\}$, the output is modeled as:
\begin{equation}
p(\mathbf{y}_d | \mathbf{x}) = \mathcal{N}(\mu_d(\mathbf{x}), \sigma_d^2 \mathbf{I}),
\end{equation}
where $\sigma_d^2$ captures domain-level uncertainty and is learned during training.

This leads to the following total loss:
\begin{equation}
\mathcal{L}_{\text{total}} = \frac{1}{2\sigma_f^2} \mathcal{L}_f + \frac{1}{2\sigma_s^2} \mathcal{L}_s + \frac{1}{2} \log \sigma_f^2 + \frac{1}{2} \log \sigma_s^2,
\end{equation}
where $\mathcal{L}_f$ and $\mathcal{L}_s$ denote the mean squared error in the fluid and solid regions, respectively.

We refer to this adaptive formulation as \textbf{IGBL}. Unlike general-purpose multi-task loss weighting~\cite{kendall2018multi, liu2019loss}, IGBL is tailored for coupled multi-physics systems. Here, $\sigma_d^2$ captures the physical complexity and learning difficulty of each domain, enabling the model to dynamically reweight losses based on predictive uncertainty. This not only stabilizes training but also improves accuracy across domains without manual tuning.

\section{Experiments}
\subsection{Dataset}
We evaluate models on two newly constructed FSI benchmarks and one public dataset. For the benchmarks, data is partitioned into training, validation, and test sets in an 8:1:1 ratio.

\subsubsection{FI-Valve}
The FI-Valve benchmark captures \textbf{F}luid-\textbf{I}nduced deformation of cardiovascular \textbf{Valve} leaflets under pulsatile inflow~\cite{bornemann2024instability}. It features tightly coupled FSI at transitional-to-high Reynolds numbers. The dataset comprises 320 simulations ($\sim$6{,}000 mesh elements each) with diverse valve geometries and inflow waveforms.

\subsubsection{SI-Vessel}
The SI-Vessel benchmark focuses on \textbf{S}tructure-\textbf{I}nduced flow variation in compliant \textbf{Vessels}. It includes both rigid and elastic wall segments subjected to time-varying pressure loading, where wall deformation alters downstream fluid dynamics~\cite{heil2011fluid}. The dataset contains 200 simulations ($\sim$13{,}000 mesh elements each), with varied geometries, material properties, and load profiles.

\subsubsection{NS+EW}
The public NS+EW dataset~\cite{rahman2024pretraining} simulates incompressible Newtonian flow past a fixed cylinder with an elastic strap in a two-dimensional channel. Following their protocol, we evaluate few-shot performance with 5, 25, and 100 training samples, using a fixed geometry and Reynolds numbers of 400 and 4000.
\subsection{Baselines}
We compare HGATSolver with a diverse set of baselines, including U-Net~\cite{ronneberger2015u} as a structured CNN-based model, GCN~\cite{kipf2017semisupervised} and GAT~\cite{veličković2018graph} as classical graph neural networks, and HGAT~\cite{wang2019heterogeneous} as a standard heterogeneous GNN. We further include GINO~\cite{NEURIPS2023_70518ea4} and GNOT~\cite{hao2023gnot} as recent neural operators for PDE learning, along with attention-based solvers Transolver~\cite{pmlr-v235-wu24r} and AMG~\cite{AMG2025}.

\subsection{Evaluation metrics}
We use the mean relative $\ell_2$ error as the primary evaluation metric. For FI-Valve and SI-Vessel, we separately evaluate fluid variables (velocity, pressure) and solid variables (displacement, stress). For the NS+EW dataset, we follow its original protocol and report a single combined error. Given $N$ samples, the mean relative $\ell_2$ error is computed as:
\begin{equation}
\text{Relative } \ell_2 = \frac{1}{N} \sum_{i=1}^{N} \frac{\left\| \hat{\mathbf{y}}^{(i)} - \mathbf{y}^{(i)} \right\|_2}{\left\| \mathbf{y}^{(i)} \right\|_2}.
\label{eq:relative_l2_mean}
\end{equation}

\subsection{Implementation Details}

For fairness, all models are implemented in PyTorch and trained using the AdamW optimizer with a batch size of 16, a temporal window of 10, and cosine learning rate scheduling for 500 epochs without early stopping. All experiments are conducted on two NVIDIA RTX 5090 GPUs. To ensure reproducibility, all models are trained using the same fixed random seed.

\subsection{Main Results}
We evaluate HGATSolver against a suite of strong baselines on our two proposed benchmarks and a public dataset to demonstrate its effectiveness.

\begin{table}[t]
\centering
\small
\setlength{\tabcolsep}{9pt}

\begin{tabular}{@{}lcccc@{}}
\toprule[1pt]
\multirow{2}{*}{Model} & \multicolumn{2}{c}{FI-Valve} & \multicolumn{2}{c}{SI-Vessel} \\
\cmidrule(lr){2-3} \cmidrule(lr){4-5}
& Fluid & Solid & Fluid & Solid \\
\midrule
U-Net (2015)        & 7.346 & 3.521 & 13.180 & 5.881 \\
GCN (2017)         & 5.486 & 0.471 & 7.204  & 0.799 \\
GAT (2018)         & 4.125 & 0.459 & 7.037  & 0.812 \\
HGAT (2019)        & 3.218 & 0.491 & 5.207  & 0.818 \\
GINO (2023)        & 3.312 & 0.489 & 5.203  & 0.827 \\
GNOT (2023)        & 3.361 & 0.402 & 6.902  & 0.768 \\
Transolver (2024)  & \underline{2.978} & 0.318 & 4.807 & \underline{0.679} \\
AMG (2025)         & 3.042 & \underline{0.312} & \underline{4.749} & 0.688 \\
\midrule
HGATSolver (Ours)  & \textbf{2.649} & \textbf{0.250} & \textbf{4.569} & \textbf{0.652} \\
\bottomrule[1pt]
\end{tabular}
\caption{Comparison of HGATSolver with baseline methods on two benchmarks. We report the mean Relative $\ell_2$ Error (\%) for both fluid and solid domains.}
\label{tab:comparison_methods}
\end{table}

\subsubsection{Quantitative Analysis in Benchmarks}
To evaluate performance across the two benchmarks, we report separate fluid and solid errors on FI-Valve and SI-Vessel in Table~\ref{tab:comparison_methods}. HGATSolver consistently achieves the lowest relative $\ell_2$ errors across all tasks. On FI-Valve, it reports 2.649\% in the fluid domain and 0.250\% in the solid domain, outperforming Transolver (2.978\%, 0.318\%) and AMG (3.042\%, 0.312\%) by up to 19.9\%. On SI-Vessel, HGATSolver achieves 4.569\% (fluid) and 0.652\% (solid), again improving over AMG (4.749\%, 0.688\%). While attention-based baselines such as Transolver and AMG perform competitively, especially in fluid regions, they do not achieve consistent accuracy across both domains. GNNs and neural operators like GINO and GNOT exhibit higher errors, particularly in fluid-heavy scenarios. HGATSolver's improvements are most evident near fluid–solid interfaces, where its heterogeneous graph design and relation-aware attention enable more accurate modeling of coupled dynamics. These results highlight the benefit of embedding domain-specific structure directly into the model architecture.

\subsection{Few-Shot Generalization Analysis}

Table~\ref{tab:l2_loss_finalstyle} highlights HGATSolver’s superior sample efficiency and robustness across flow regimes.  It achieves 0.237\% error with only 5 training samples at $Re=400$, outperforming all other supervised methods and indicating strong generalization even in severely limited data regimes. With 100 samples, the error drops to 0.055\%, significantly lower than the next best model at 0.109\%. At $Re=4000$, where nonlinear flow behavior increases learning difficulty, HGATSolver continues to lead with 0.270\% error, reflecting stable learning across scales. In contrast, conventional GNNs and attention-based baselines exhibit large error gaps between low and high Reynolds settings or saturate early with increased data. These trends underscore that the model’s heterogeneous architecture and adaptive gating are not only effective for encoding physical inductive biases but also critical for maintaining accuracy and consistency in sparse data and complex regimes.
\begin{table}[htbp]
\centering
\small
\setlength\tabcolsep{3pt}
\begin{tabular}{lcccccc}
\toprule
\multirow{3}{*}{Model}
& \multicolumn{3}{c}{$Re = 400$} 
  & \multicolumn{3}{c}{$Re = 4000$} \\
\cmidrule(r){2-4} \cmidrule(l){5-7}
  & \multicolumn{6}{c}{Training Samples} \\
  & 5 & 25 & 100 & 5 & 25 & 100 \\
\midrule
U-Net (2015)         & 1.891 & 0.917 & 0.451 & 1.805 & 0.750 & 0.540 \\
GCN (2017)          & 0.291 & 0.195 & 0.164 & 0.767 & 0.497 & 0.333 \\
GAT (2018)          & 0.313 & 0.212 & 0.179 & 0.809 & 0.518 & 0.354 \\
HGAT (2019)         & \underline{0.279} & 0.187 & 0.158 & \underline{0.730} & 0.491 & 0.321 \\
GINO (2023)         & 0.349 & 0.230 & 0.207 & 0.846 & 0.540 & 0.369 \\
Transolver (2024)   & 0.324 & 0.139 & \underline{0.109} & 0.769 & 0.493 & \underline{0.309} \\
AMG (2025)   & 0.302 & \underline{0.132} & 0.130 & 0.798& \underline{0.490} & 0.343 \\
\midrule
HGATSolver (Ours)   & \textbf{0.237} & \textbf{0.084} & \textbf{0.055} & \textbf{0.540} & \textbf{0.475} & \textbf{0.270} \\
\bottomrule
\end{tabular}
\caption{Relative $\ell_2$ Error (\%) on the NS+EW dataset with varying Reynolds numbers and training sample sizes.}
\label{tab:l2_loss_finalstyle}
\end{table}

\subsection{Qualitative Error Visualization}
\begin{figure}[t]
\centering
\includegraphics[width=1.0\columnwidth]{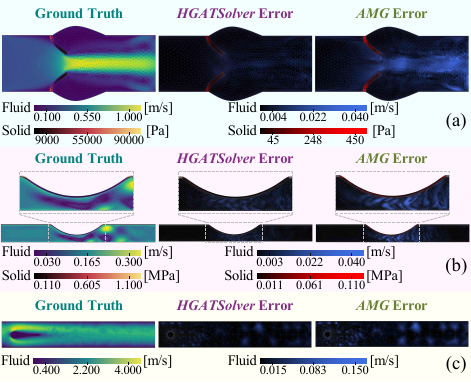}
\caption{Prediction error maps for HGATSolver and AMG across three datasets: (a) FI-Valve, (b) SI-Vessel, and (c) NS+EW.}
\label{fig:qualitative_results}
\end{figure}

The spatial distribution of prediction errors across three cases is illustrated in Fig.~\ref{fig:qualitative_results}. In (a) FI-Valve, HGATSolver sharply reduces error in the valve leaflet and surrounding flow, particularly near the fluid–solid interface where the AMG accumulates high residuals. In (b) SI-Vessel, both fluid pressure and solid stress fields are better captured, especially across elastic–rigid material junctions, indicating that HGATSolver adapts well to varying material properties. In (c) NS+EW, HGATSolver suppresses error in high-velocity fluid jets, which the AMG fails to resolve. These results confirm HGATSolver's ability to model multi-domain dynamics more precisely, especially in regions of strong coupling or gradient shifts.

\subsection{Ablation study}
We evaluate the contribution of HGATSolver’s main components through targeted ablations.

\subsubsection{Quantitative Impact of Components}
We assess the contribution of each architectural module through an ablation study on FI-Valve and SI-Vessel, with results shown in Table~\ref{tab:ablation_custom}. Removing the PCGM (w/o PCGM) leads to the largest performance drop, with fluid error on FI-Valve increasing from 2.649\% to 3.276\%, and solid error on SI-Vessel rising from 0.652\% to 0.851\%. This highlights its role in stabilizing message updates across fluid–solid transitions. Disabling the IGBL (w/o IGBL) results in notable error increases in solid regions (e.g., 0.250\% $\rightarrow$ 0.292\% on FI-Valve), suggesting the importance of adaptive objective weighting for coupled systems. The removal of time embeddings (w/o Time Embedding) degrades performance across both domains, particularly in FI-Valve solids (0.250\% $\rightarrow$ 0.315\%), indicating the benefit of temporal conditioning. Learnable aggregation (w/o Learnable Agg.) and physics parameters (w/o Physics Params.) also contribute measurably, though to a lesser extent.
\begin{table}
\centering
\small
\setlength{\tabcolsep}{7pt}
\begin{tabular}{@{}lcccc@{}}
\toprule[1pt]
\multirow{2}{*}{\begin{tabular}[c]{@{}l@{}}Model\\ Configuration\end{tabular}} & \multicolumn{2}{c}{FI-Valve} & \multicolumn{2}{c}{SI-Vessel} \\
\cmidrule(lr){2-3} \cmidrule(lr){4-5}
& \begin{tabular}[c]{@{}c@{}}Fluid\end{tabular} & \begin{tabular}[c]{@{}c@{}}Solid\end{tabular} & \begin{tabular}[c]{@{}c@{}}Fluid \end{tabular} & \begin{tabular}[c]{@{}c@{}}Solid \end{tabular} \\
\midrule
Full Model & 2.649 & 0.250 & 4.569 & 0.652 \\
\midrule
w/o Physics Params. & 2.721 & 0.262 & 4.873 & 0.735 \\
w/o PCGM & 3.276 & 0.340 & 5.284 & 0.851 \\
w/o Learnable Agg. & 3.054 & 0.287 & 4.891 & 0.699 \\
w/o IGBL & 2.853 & 0.292 & 4.793 & 0.711 \\
w/o Time Embedding & 3.119 & 0.315 & 4.782 & 0.678 \\
\bottomrule[1pt]
\end{tabular}
\caption{Ablation study of HGATSolver. We report the mean Relative $\ell_2$ Error (\%) on two benchmarks.}
\label{tab:ablation_custom}
\end{table}

\begin{figure}[t]
\centering
\includegraphics[width=1.0\columnwidth]{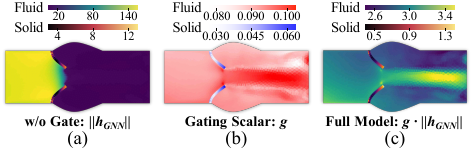}
\caption{Visualization on FI-Valve: (a) GNN output magnitude without gating, (b) learned physics-conditioned gating scalar $g$, and (c) final effective update $g \cdot ||h_{\text{GNN}}||$.}
\label{fig:gating_mechanism}
\end{figure}
\subsubsection{Analyzing the PCGM}

We examine how the PCGM regulates update magnitudes across heterogeneous domains. As shown in Fig.~\ref{fig:gating_mechanism}(a), the ungated GNN outputs ($||h_{\text{GNN}}||$) exhibit high and spatially inconsistent magnitudes in both fluid and solid regions, particularly near deforming vessel walls. The learned gating scalar $g$ in Fig.~\ref{fig:gating_mechanism}(b) suppresses updates in stiffer solid regions while retaining larger values in dynamically active fluid zones such as central jets. The gated output in Fig.~\ref{fig:gating_mechanism}(c), computed as $g \cdot ||h_{\text{GNN}}||$, reflects a controlled update profile that preserves necessary nonlinearity in fluid regions while attenuating overshooting in solids.

This modulation contributes directly to the model’s stability and accuracy. By conditioning updates on both physical priors and learned features, PCGM reduces excessive variations near interfaces and stiff materials, thereby preventing instabilities in the state update. The visualized differences confirm that PCGM enables more consistent behavior across domains, validating its role in improving numerical robustness in coupled dynamics.

\subsubsection{Effectiveness of IGBL}

\begin{figure}[t]
\centering
\includegraphics[width=1.0\columnwidth]{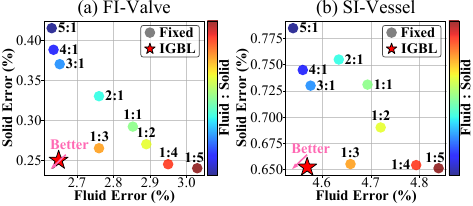}
\caption{Effectiveness of the IGBL on the (a) FI-Valve and (b) SI-Vessel. Colored points represent models trained with fixed, manually tuned fluid/solid loss weights. The red star indicates HGATSolver trained with IGBL.}
\label{fig:igbl_effectiveness}
\end{figure}

We evaluate the IGBL by comparing it to fixed-weight baselines using manually chosen fluid-to-solid loss ratios. For each benchmark, we sweep a range of static weightings (e.g., 1:1 to 1:5) and record the relative $\ell_2$ errors in both fluid and solid domains. As shown in Fig.~\ref{fig:igbl_effectiveness}, the results form a smooth Pareto front, indicating an inherent trade-off: reducing error in one domain often increases it in the other. On FI-Valve, for example, a 1:3 ratio improves solid accuracy but raises fluid error, while more balanced ratios yield mediocre results in both domains.

In contrast, HGATSolver with IGBL automatically learns an effective balance without manual tuning. The resulting solution lies strictly below the Pareto front in both benchmarks, achieving lower errors in both domains simultaneously. This outcome is not interpolated from fixed-weight points but emerges from IGBL’s adaptive reweighting, which adjusts gradient contributions based on per-domain predictive uncertainty. The decoupling of domain sensitivities during training allows optimization to avoid trade-off-dominated regimes and converge toward a more favorable solution. These results directly validate that IGBL not only improves final accuracy but also simplifies model selection in multi-objective coupled learning tasks.

\section{Conclusion}
We propose \textbf{HGATSolver}, a learning-based solver for coupled FSI systems that leverages a heterogeneous graph architecture to encode domain-specific physics. Combined with \textbf{PCGM} for enhanced stability and \textbf{IGBL} for adaptive training, HGATSolver achieves state-of-the-art accuracy on two challenging FSI benchmarks we constructed, and shows strong few-shot generalization on a public dataset. Ablation studies confirm that each component contributes independently and complementarily to performance. These results establish HGATSolver as a principled and effective framework for surrogate modeling of multi-physics systems.
\section{Acknowledgments}
This work was supported by the National Natural Science Foundation of China under Grant 62373352.

\bibliography{aaai2026}


\end{document}